\documentclass[11pt]{article}

\usepackage[final]{acl}

\usepackage{times}
\usepackage{latexsym}

\usepackage[T1]{fontenc}

\usepackage[utf8]{inputenc}

\usepackage{microtype}

\usepackage{inconsolata}

\usepackage{graphicx}
\usepackage{amsmath}
\usepackage{amssymb}
\usepackage{amsthm}
\usepackage{booktabs}
\usepackage[english]{babel}
\usepackage{makecell}
\usepackage{multirow} 
\usepackage{wrapfig}
\usepackage{subcaption}
\usepackage[ruled,linesnumbered]{algorithm2e}
\usepackage{pifont}
\usepackage[table]{xcolor}
\newcommand{\cco}[1]{\cellcolor{orange!25}#1}  
\newcommand{\ccr}[1]{\cellcolor{red!25}#1}     
\usepackage[most]{tcolorbox}
\title{Cognitive Loop of Thought: Reversible Hierarchical Markov Chain for Efficient Mathematical Reasoning}

\author{ 
\textbf{Jia-Chen Zhang\textsuperscript{1}\footnotemark[2]},   
\textbf{Yu-Jie Xiong\textsuperscript{2}\footnotemark[1]}, 
\textbf{Zheng Zhou\textsuperscript{2}\footnotemark[2]}, 
\\ 
\textsuperscript{1}School of Computer Science and Technology, East China Normal University \\ 
3663 North Zhongshan Road, Shanghai, China\\
\textsuperscript{2}School of Electronic and Electrical Engineering, Shanghai University of Engineering Science \\ 
333 Longteng Road, Shanghai, China
\\ 
\small{   
\textbf{Correspondence:}    
\href{mailto:xiong@sues.edu.cn}{xiong@sues.edu.cn},    
}
}

\begin{document}
\maketitle
\begin{abstract}
Multi-step Chain-of-Thought (CoT) has significantly advanced the mathematical reasoning capabilities of LLMs by leveraging explicit reasoning steps. However, the widespread adoption of Long CoT often results in sequence lengths that exceed manageable computational limits. While existing approaches attempt to alleviate this by reducing KV Cache redundancy via Markov chain-like structures, they introduce two critical limitations: inherent memorylessness (loss of context) and limited backward reasoning capability.
To address these limitations, we propose a novel Chain-of-Thought framework based on Reversible Hierarchical Markov Chain, termed Cognitive Loop of Thought (CLoT), and a backward reasoning dataset CLoT-Instruct. In CLoT, problems are decomposed into sub-problems with hierarchical dependencies. Inspired by human cognitive processes—where reasoning is revisited to verify conclusions—we introduce a backward verification mechanism at each hierarchical layer. Furthermore, we implement a pruning strategy: once higher-level sub-problems are verified, redundant lower-level sub-problems are pruned to maximize efficiency. This approach effectively mitigates error propagation and enhances reasoning robustness. Experiments on four mathematical benchmarks demonstrate the effectiveness of our method. Notably, on the AddSub dataset using gpt-4o-mini, CLoT achieves 99.0\% accuracy, outperforming traditional CoT and CoT-SC by 4.1\% and 2.9\%, respectively. Our code is publicly available at: \url{https://anonymous.4open.science/r/CLoT-7EBD}.
\end{abstract}

\section{Introduction}
Large Language Models (LLMs) have achieved remarkable breakthroughs across a broad spectrum of Natural Language Processing (NLP) tasks, such as question answering, automatic summarization, and machine translation \cite{achiam2023gpt,chowdhery2023palm,touvron2023llama,huang-etal-2023-large, zhao2023survey}. Despite these achievements, LLMs continue to exhibit notable limitations in reasoning performance when compared to human cognitive capabilities. Empirical evidence suggests that merely increasing model scale through parameter expansion is insufficient to bridge the gap in complex reasoning abilities between current LLMs and human-level intelligence. \cite{zhou2024larger}.

To enhance the model's reasoning capabilities, researchers have explored various strategies to enhance the logical coherence and step-by-step inference processes of LLMs. One of the most influential approaches is Chain-of-Thought (CoT) reasoning \cite{NEURIPS2022_8bb0d291}, which enables models to generate intermediate reasoning steps before arriving at a final answer. By mimicking human-like problem-solving patterns, CoT not only improves model performance on complex reasoning tasks, but also enhances the interpretability of model decisions. The transparency afforded by explicit reasoning traces allows users to inspect, validate, and potentially correct the model’s logic, thereby fostering trust and facilitating debugging.
\begin{figure}[t]
  \centering
  \includegraphics[width=1.0\linewidth]{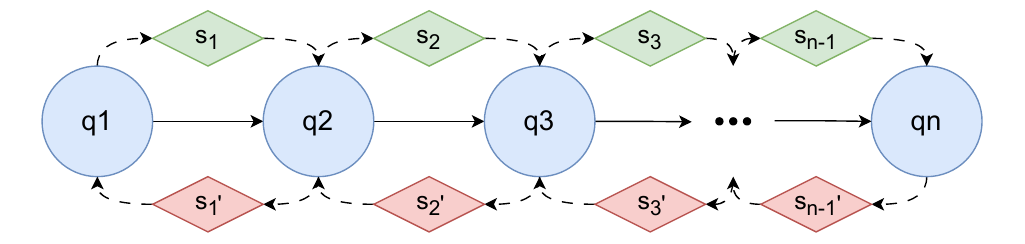}
  \caption{Overview of the Reversible Hierarchical Markov Chain Framework.}
  \label{photoup}
\end{figure}

However, when facing real-world application scenarios, CoT still encounters the challenge that intermediate reasoning steps may contain errors. To address this challenge, one approach to improve performance is intrinsic self-correction \cite{10.1162/tacl_a_00713,10.1162/tacl_a_00660}, which allows the model to check and revise its own generated answers without external feedback—a process highly analogous to human thinking. Nevertheless, negative views on self-correction also exist. \cite{stechly2023gpt, tyen-etal-2024-llms, jiang2024self} find that large language models are even unable to determine the correctness of answers, often changing correct answers into incorrect ones or failing to correct originally erroneous responses. The debates in prior research indicate that the self-correction mechanisms of large models still lack in-depth exploration.

To address these limitations and obviate the need for extensive training on verification tasks, we propose a novel reverse self-verification method named CLoT (Cognitive Loop of Thought). Humans often employ "reverse engineering" or self-validation to confirm inferred conclusions. Drawing inspiration from this cognitive process and reversible Markov chains, we treat the forward refinement and backward verification steps as a single closed loop, iteratively executing this process to ensure the validity of intermediate reasoning. Specifically, in the forward refinement phase, the LLM progressively generates sub-questions and answers via CoT. Subsequently, each sub-question undergoes backward verification: the original conditions are treated as unknown variables, while the generated answer serves as a known condition to deduce the original premises. Consistency between the backward inference and the original condition validates the step; discrepancies trigger a root-cause analysis. This bidirectional verification minimizes error propagation in the reasoning chain. We evaluate our method using gpt-4o-mini, gpt-4-1106-preview and DeepSeek-V3 across multiple mathematical, commonsense, and logical reasoning datasets. Our contributions are summarized as follows:
\begin{itemize}
\item[$\bullet$] We propose a novel self-reflective verification method for reasoning chains, named CLoT (Cognitive Loop of Thought). CLoT is capable of verifying both intermediate steps and the final answer, as well as analyzing the causes of errors.
\end{itemize}
\begin{itemize}
\item[$\bullet$] We introduce a hierarchical pruning strategy that removes redundant verification steps at lower layers, reducing CLoT's resource consumption by 41.8\% without compromising performance.
\end{itemize}
\begin{itemize}
\item[$\bullet$] We construct CLoT-Instruct, an instruction-tuning dataset designed to facilitate the learning of "Backward Verification" capabilities in LLMs.
\end{itemize}
\begin{itemize}
\item[$\bullet$] Extensive experiments across six diverse tasks demonstrate that CLoT effectively identifies and corrects reasoning errors, consistently outperforming baseline methods.
\end{itemize}

\begin{figure*}[t]
  \centering
  \includegraphics[width=1.0\linewidth]{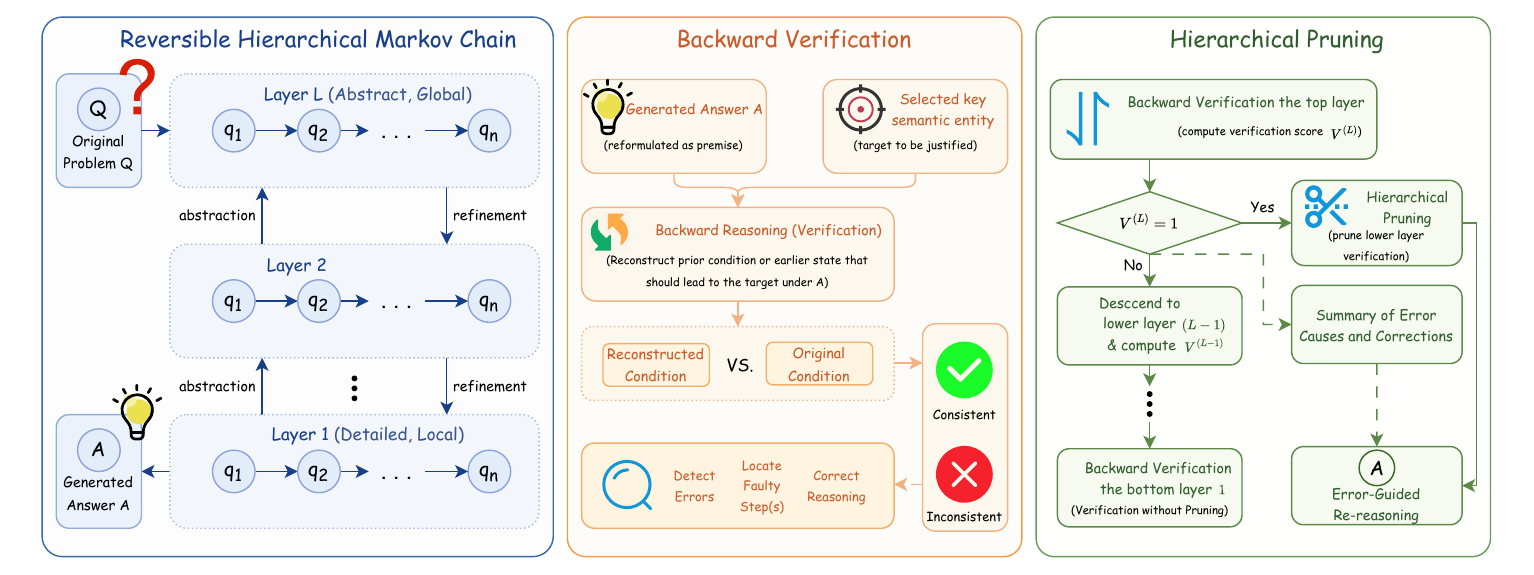}
  \caption{Overview of the CLoT Framework.}
  \label{photoup}
\end{figure*}

\section{Related Work}
Since training LLMs requires enormous resources, effectively improving model performance without retraining has become a key research focus. \cite{NEURIPS2022_9d560961} introduced the concept of CoT reasoning, emphasizing the importance of deriving conclusive answers through multi-step logical pathways. \cite{NEURIPS2022_8bb0d291} found that simply adding the phrase "let's think step by step" to prompts enables LLMs to perform zero-shot logical reasoning without any additional human-designed prompts. Subsequently, \cite{wang2023selfconsistency} proposed Self-Consistency (SC) to replace the greedy decoding strategy. \cite{zhang2023automatic} developed an automatic CoT framework based on the input problem, eliminating the instability caused by manual prompting. \cite{fu2023complexitybased} employed complexity-based multi-step reasoning estimation to execute CoT. \cite{zhu2025understanding} further enhanced performance by applying pre-prompting in the form of a plugin. However, standard chain-like reasoning still struggles with highly complex logical problems. Variants such as Tree-of-Thoughts \cite{NEURIPS2023_271db992}, Graph-of-Thoughts \cite{besta2024graph}, and Markov-chain-inspired reasoning frameworks have effectively enhanced the model's reasoning capabilities \cite{teng2025atom, yang2025markov}. Nevertheless, in practical applications, large models often fail to recognize their own errors and tend to repeat mistakes, limiting their applicability in real-world scenarios.

To expand the application scenarios of LLMs by enhancing their ability to identify and correct errors, researchers have developed various methods to improve their self-correction capabilities. FOBAR \cite{jiang-etal-2024-forward} relies on Self-Consistency, requiring multiple direct reasoning attempts (e.g., 10 times) for the same problem and triggering verification only when answers are inconsistent. \cite{wang2023shepherd} verifies answers by integrating inputs from both humans and other models. \cite{weng-etal-2023-large} employs backward verification to evaluate and score multiple candidate answers generated during a tree-of-thought reasoning process. \cite{li2024confidence} enables models to effectively distinguish between confident and uncertain responses using a simple "Self-Verifying" prompt (e.g., "If you are very confident about your answer, maintain your answer. Otherwise, update your answer"). \cite{zheng2024progressivehint} relies on ground truth to determine if a question needs to be answered again. \cite{10.5555/3692070.3692341} identifies and corrects errors by having the LLMs analyze the intermediate steps of a CoT for mistakes and pinpoint the source of the error.
While these methods have improved self-correction to varying degrees, they often face the challenges summarized by \cite{yang2024confidence}: (1) reliance on high-quality external feedback, (2) lack of comparison with baselines that consume the same number of tokens, and (3) prompts that are strictly designed and potentially complex. 

To address these challenges, we propose the CLoT. Our approach achieves self-correction without requiring any external high-quality feedback. It demonstrates a significant performance improvement compared to a baseline with equivalent token consumption and operates in a zero-shot manner, eliminating the need for designing complex prompt examples.

\section{Method}
In this section, we provide a detailed description of the design of the CLoT method. CLoT guides the model to perform backward verification, identifies potential failures in backward verification, and ultimately summarizes the causes of errors and corrects them. Additionally, we construct a CLoT-Instruct Dataset based on the CLoT approach to help train models to learn "Backward Verification" approach.
\subsection{Reversible Hierarchical Markov Chain}
For a mathematical problem $q$, successful reasoning involves not only the sequential simplification of the problem through derivation steps but also an implicit hierarchical organization of cognitive processes---such as conceptual abstraction, subproblem decomposition, and strategic planning. To capture this structure, we propose a Hierarchical Markov Chain framework that models reasoning as a multi-level stochastic process, where each level corresponds to a distinct granularity of problem understanding or solution strategy.

The architecture of CLoT is illustrated in Figure \ref{photoup}, where the left panel delineates the structure of the Reversible Hierarchical Markov Chain. Specifically, CLoT first decomposes the target problem into a Markov chain of intermediate subproblems and explicitly outputs the logical dependencies among them. Subsequently, the model stratifies this sequence into discrete reasoning layers $l \in \{1,\dots,L\}$ based on dependency counts. Layer $L$ encompasses fewer dependencies and represents the most abstract strategy level, where the original problem is instantiated as the root node $q^{(L)}$. Conversely, layer $1$ contains the most dependencies, representing the finest-grained derivation level. This cross-layer process of abstraction and refinement is probabilistically modeled as inter-layer propagation:
\begin{equation}
p(q^{(l+1)} \mid q^{(l)}) \quad \text{(abstraction)},
\label{eq:1}
\end{equation}
\begin{equation}
p(q^{(l)} \mid q^{(l+1)}) \quad \text{(refinement)}.
\label{eq:2}
\end{equation}

Within each layer $l$, the reasoning process unfolds into an intra-layer Markov chain of subproblems $\{q_t^{(l)}\}_{t \ge 1}$, seeded by the layer's root representation $q_1^{(l)}$. Given the current subproblem $q_t^{(l)}$ and its corresponding chain-of-thought reasoning step $s_t^{(l)}$ that resolves it, the transition to the next subproblem satisfies the Markov property:
\begin{equation}
p(q_{t+1}^{(l)} \mid q_t^{(l)}, s_t^{(l)}).
\label{eq:0}
\end{equation}

However, this hierarchical organization alone provides only a structural skeleton; it does not inherently guarantee the logical validity of each local derivation step. To address this and prevent error accumulation, we embed a reversible verification mechanism that operates layer by layer and subproblem by subproblem. To initiate this backward verification, the model-generated final answer is reformulated as an initial premise. The model then traverses the reasoning trajectory in reverse. At each step, rather than simply relying on forward generation, we introduce two complementary directional checks:
\begin{itemize}
    \item \textbf{Forward Deduction}: The model derives the next state $q_{t+1}^{(l)}$ from $q_t^{(l)}$ via reasoning step $s_t^{(l)}$, representing standard local decision-making.
    \item \textbf{Backward Justification}: Given the consequent state $q_{t+1}^{(l)}$, the model verifies whether the antecedent state $q_t^{(l)}$ can be semantically and logically reconstructed. This serves as a semantic backtracking mechanism that asks: \textit{Is the premise justifiable given the conclusion?}
\end{itemize}

A valid reasoning step must satisfy both directional validations. Through this step-by-step backward justification, the model ultimately attempts to reconstruct specific conditions from the original problem (or identify key semantic entities for non-numerical tasks). By determining whether these reconstructed target variables align with the original input, CLoT completes its dual-loop verification. This approach provides a fine-grained diagnostic signal to locate and rectify reasoning flaws or non sequiturs without requiring external supervision, parameter updates, or score-based candidate selection, thereby substantially enhancing the model's interpretability, safety, and reliability.

\subsection{Hierarchical Pruning}
Building upon the RHMC framework, we introduce a Consistency-Aware Hierarchical Pruning strategy. This method exploits hierarchical dependencies to minimize computational redundancy without sacrificing verification rigor. The core premise is that global semantic consistency at higher abstraction layers is a high-fidelity proxy for lower-level correctness. In CLoT’s backward reconstruction, if all sub-questions within the highest layer $L$ are successfully verified, the entire reasoning trajectory forms a closed logical loop. Formally, for a layer $l$ with $T_l$ sub-questions, we define the layer-wise verification status $\mathcal{V}^{(l)} \in \{0, 1\}$ as the logical conjunction of all individual sub-question outcomes:
\begin{equation}
\mathcal{V}^{(l)} = \prod_{t=1}^{T_l} \mathbb{I} \Big( \text{Verify}(q_t^{(l)} \mid q_{t+1}^{(l)}) \Big),
\end{equation}
where $\mathbb{I}(\cdot)$ is an indicator function returning $1$ if the backward reconstruction of $q_t^{(l)}$ is strictly consistent, and $0$ otherwise. This boolean gatekeeper drives a top-down dynamic pruning protocol:
\begin{itemize}
\item \textbf{Macro-Verification} \& Pruning: The system evaluates from the top layer $L$. A perfect verification ($\mathcal{V}^{(l)} = 1$) indicates the final output is logically anchored to the input. Once confirmed, the system immediately prunes all lower-layer verifications, trusting that a reconstructible "big picture" implicitly validates fine-grained derivations.

\item \textbf{Recursive Refinement}: If $\mathcal{V}^{(l)} = 0$, it signals a potential hallucination or logical gap. The system then descends to layer $l-1$ for granular validation. This operates recursively: the model either halts and prunes upon achieving $\mathcal{V}^{(l)} = 1$ at any level, or descends until it pinpoints a specific failure at the finest-grained layer.
\end{itemize}
Crucially, this pruning is outcome-dependent. Subtle low-level errors that corrupt the final answer will inevitably cause high-level backward verification to fail. This automatically breaks the $\mathcal{V}^{(l)} = 1$ condition, forcing a mandatory descent. Thus, hierarchical pruning is triggered only when the reasoning is flawless across the entire semantic scope.

\subsection{CLoT-Instruct Dataset}
\subsubsection{Dataset Construction}
Building upon the theoretical framework of the RHMC, we introduce CLoT-Instruct, a specialized instruction-tuning dataset designed to train language models with bidirectional reasoning and self-verification capabilities. Unlike conventional chain-of-thought (CoT) datasets that focus solely on forward reasoning, CLoT-Instruct explicitly encodes the backward verification mechanism central to our CLoT framework, enabling models to develop intrinsic self-consistency checking abilities.

Theoretically, our construction methodology is applicable to any mathematical reasoning dataset, as it does not rely on specific problem types but instead builds upon a general hierarchical reasoning structure and bidirectional verification logic. Currently, we have implemented and released three high-quality subsets based on this approach, using GSM8K, SVAMP, and AddSub as foundational datasets. The construction of CLoT-Instruct follows a two-stage pipeline designed to jointly ensure the quality of forward reasoning and the integrity of backward verification:
\begin{itemize}
    \item \textbf{Forward Reasoning}: Creation of hierarchical problem-solving trajectories.
    \item \textbf{Backward Verification}: Generation of logically inverted validation questions.
\end{itemize}

\subsubsection{Dataset Structure}
\label{sec:dataset_structure}
Formally, let $\mathcal{D}_{\text{origin}}$ denote the original mathematical reasoning dataset, where each instance follows the format $\tau = (q, s_{1:T}, a)$. Here, $q$ represents the initial problem, $s_{1:T}$ denotes a sequence of $T$ reasoning steps ($T \geq 1$), and $a$ is the final answer.

In contrast to conventional approaches that treat the entire reasoning chain as a monolithic unit, our CLoT framework explicitly decomposes reasoning into hierarchical components, each augmented with bidirectional verification signals. To support this, we construct the CLoT-Instruct dataset, in which every sample is uniformly structured as:
\begin{equation}
    \mathcal{X} = \Bigl(
        q_{\text{origin}},\;
        a_{\text{gt}},\;
        L,\;
        \{ \tau^{(l)} \}_{l=1}^{L}
    \Bigr),
    \label{eq:dataset_structure}
\end{equation}
where $q_{\text{origin}}$ is the original problem, $a_{\text{gt}}$ is its ground-truth answer, and $L$ denotes the maximum reasoning depth required for this specific instance. For each reasoning layer $l \in \{1, \dots, L\}$, the trajectory $\tau^{(l)}$ is defined as:
\begin{equation}
    \tau^{(l)} = \left( q^{(l)},\; s^{(l)}_{1:T_l},\; \bigl\{ (q_{\text{verify},t}^{(l)},\, a_{\text{verify},t}^{(l)}) \bigr\}_{t=1}^{T_l} \right),
\end{equation}
where $q^{(l)}$ is the problem representation at layer $l$ (either the original question or an abstracted sub-question), and $s^{(l)}_{1:T_l}$ is its corresponding sequence of reasoning steps. Crucially, each pair $(q_{\text{verify},t}^{(l)}, a_{\text{verify},t}^{(l)})$ constitutes a bidirectional verification instance: $q_{\text{verify},t}^{(l)}$ is a logically inverted question derived from the forward step, and $a_{\text{verify},t}^{(l)}$ is its expected answer (representing the reconstructed antecedent). 

This unified representation enables hierarchical dependency tracking, dynamic verification, and scalable supervision across various reasoning depths, all while maintaining a consistent and extensible data schema.

\begin{table*}
\small
\centering
\tabcolsep=0.38cm
\renewcommand\arraystretch{1.0}
\begin{tabular}{l|cccccc|c}
\toprule
\textbf{Models} & \textbf{AddSub} & \textbf{GSM8K} & \textbf{SVAMP} & \textbf{MATH} & \textbf{AQuA} & \textbf{CSQA} & \textbf{Avg.}\\
\midrule
\textbf{gpt-4o-mini}\\
\midrule
CoT &94.9 &90.9 &93.4 &78.3&81.4&81.5&86.7\\
CoT-SC(n=5)&96.1&92.0&93.5&\ccr{81.0}&83.1&\ccr{83.4}&\cco{88.3}\\
ToT&76.7&91.5&93.2&80.7&72.3&81.7&82.7\\
AR&82.3&91.7&92.4&73.1&77.5&78.7&81.0\\
C-CoT&\cco{97.2}&\cco{92.4}&93.4&76.9&83.3&81.0&87.4\\
ISP-2&96.1&\cco{92.4}&\cco{93.9}& 77.9&\cco{83.7}&81.9& 87.7\\
CLoT(ours) &\ccr{99.0$\pm 0.0$}&\ccr{94.6$\pm 0.2$}&\ccr{94.9$\pm 0.1$}&\cco{80.7$\pm 0.3$}& \ccr{85.8$\pm 0.0$}&\cco{82.3$\pm 0.3$}&\ccr{89.6}\\
\midrule
\textbf{gpt-4}\\
\midrule
CoT &98.0 &93.1&93.4 &73.0&82.7&84.9&87.5\\
CoT-SC(n=5)&\cco{98.4}&\cco{94.2}&93.5&\ccr{81.6}&83.1&\ccr{85.6}&\cco{89.4}\\
AR&87.1&93.4&92.7&73.1&79.5&79.1&84.2\\
C-CoT&\cco{98.4}&\cco{94.2}&\cco{93.6}&80.6&82.7&83.9&88.9\\
ISP-2&97.7&93.3&93.5&77.7&\cco{84.2}&83.4&88.3\\
Tr&97.5&\cco{94.2}&91.3&71.9&79.9&79.4&85.7\\
CLoT(ours) &\ccr{99.0$\pm 0.0$}&\ccr{95.4$\pm 0.4$}&\ccr{95.0$\pm 0.1$}&\ccr{81.6$\pm 0.6$}& \ccr{86.9$\pm 0.0$}&\cco{84.9$\pm 0.1$}&\ccr{90.5}\\
\midrule
\textbf{DeepSeek-V3}\\
\midrule
CoT &97.7 &94.1&93.9 &90.2&84.2&81.1&90.2\\
CoT-SC(n=5)&\cco{98.5}&\cco{94.5}&\cco{94.3}&\cco{91.3}&87.4&\cco{82.2}&\cco{91.4}\\
AR&93.9&93.3&93.1&89.7&86.6&81.0&89.6\\
C-CoT&97.9&94.2&93.8&89.7&\cco{88.1}&81.2&90.5\\
CLoT(ours) &\ccr{99.0$\pm 0.0$}&\ccr{95.5$\pm 0.2$}&\ccr{95.6$\pm 0.2$}&\ccr{91.7$\pm 0.2$}& \ccr{91.8$\pm 0.3$}&\ccr{82.6$\pm 0.8$}&\ccr{92.7}\\
\bottomrule
\end{tabular}
\caption{Accuracy of CLoT and baselines on six mainstream datasets. Higher is better for all metrics. The \colorbox{red!25}{best score}, and \colorbox{orange!25}{second best score} are red, and orange, respectively.}
\label{main1}
\end{table*}

\section{Experiment}
In this section, we conduct comprehensive experiments to thoroughly investigate CLoT through extensive benchmark evaluations on six standard datasets, efficiency analysis, data result analysis, and ablation studies.
\subsection{Experimental Settings}
\subsubsection{Datasets}
\label{dataset-cite}
We evaluate CLoT using gpt-4o-mini, gpt-4-1106-preview and DeepSeek-V3 as backbone models. Our evaluation covers three types of reasoning tasks: mathematical QA reasoning (MATH$_{900}$, \cite{hendrycks2021measuring}, GSM8K$_{1319}$, \cite{cobbe2021training}, Addsub$_{395}$, \cite{hosseini2014learning}, and SVAMP$_{1000}$, \cite{patel-etal-2021-nlp}), mathematical multiple-choice (AQUA$_{254}$, \cite{ling-etal-2017-program}), and commonsense reasoning (CommonsenseQA$_{1220}$, \cite{talmor-etal-2019-commonsenseqa}). For the large datasets Math and SVAMP, we adopt the same data processing procedure as in Study \cite{10.5555/3692070.3692341}. On mathematical datasets, CLoT employs exact numeric matching—i.e., an answer is considered correct only if it exactly matches the numeric value in the original question. On commonsense datasets, it uses synonym-based judgment—i.e., an answer is deemed correct if it is a synonym or near-synonym of the answer in the original question.


\subsection{Main Results}
Table \ref{main1} summarizes the performance of CLoT and various baselines across six benchmarks using three different model backbones: gpt-4o-mini, gpt-4, and DeepSeek-V3. Overall, CLoT consistently demonstrates competitive performance, outperforming all competitive baselines in terms of average accuracy.

Under the gpt-4o-mini setting, CLoT achieves an average accuracy of 89.6\%, surpassing the most competitive baseline, CoT-SC (n=5), by a margin of 1.3\%. Notably, CLoT establishes new performance ceilings on AddSub (99.0\%), GSM8K (94.6\%), SVAMP (94.9\%), and AQuA (85.8\%), demonstrating its robust reasoning capabilities across both straightforward arithmetic and complex multi-step reasoning tasks. When evaluated on the more powerful gpt-4 backbone, CLoT further extends its lead, attaining the highest average accuracy of 90.5\%. Compared to the best-performing baseline, it delivers substantial gains on GSM8K (+1.2\%), SVAMP (+1.5\%), and AQuA (+3.8\%). These results underscore CLoT’s ability to effectively leverage its iterative generate-and-verify mechanism to rectify subtle errors in the reasoning chain, particularly in challenging datasets like AQuA. Furthermore, CLoT demonstrates superior scalability and generalization on DeepSeek-V3, achieving a peak average accuracy of 92.7\%. The consistent improvements across diverse tasks, ranging from symbolic arithmetic (AddSub) to complex mathematical reasoning (MATH) and CSQA, validate the effectiveness and versatility of our proposed verification driven framework.

\begin{figure}[t]
  \centering
  \includegraphics[width=1.0\linewidth]{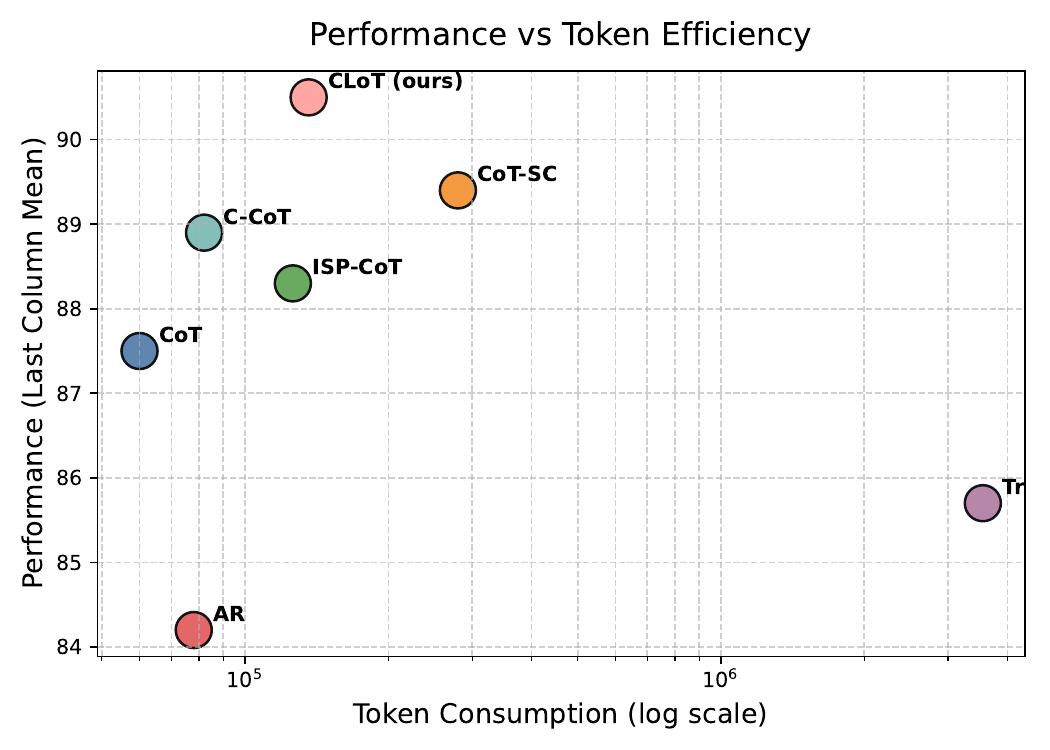}
  \caption{The results of CLoT efficiency. The token consumption corresponds to the total number of tokens used to solve the same 100 gsm8k problem.}
  \label{photoabl}
\end{figure}

\subsection{Efficiency Analysis}
We further evaluate the efficiency of different methods in terms of token consumption. The results are shown in Figure \ref{photoabl} and were obtained using the gpt-4o-mini API. Traditional methods such as CCoT and CoT-SC consume approximately 280k tokens. In contrast, Tr uses chain-of-thought for self-verification and requires as many as 3.3M tokens, resulting in very high computational costs. AR is more token-efficient, using only 78k tokens. However, its overall performance still lags behind other approaches. Notably, CLoT achieves performance on par with or even better than existing methods while consuming just 136k tokens. This highlights its strong reasoning efficiency under low token overhead. This efficiency stems from its hierarchical pruning strategy, which selectively removes redundant verification steps at lower reasoning layers. This design enables high reasoning efficacy with minimal token overhead, making CLoT particularly suitable for large-scale and resource-constrained applications.

\begin{figure*}[t]
  \centering
  \begin{subfigure}{0.49\linewidth}
    \centering
    \includegraphics[width=\linewidth]{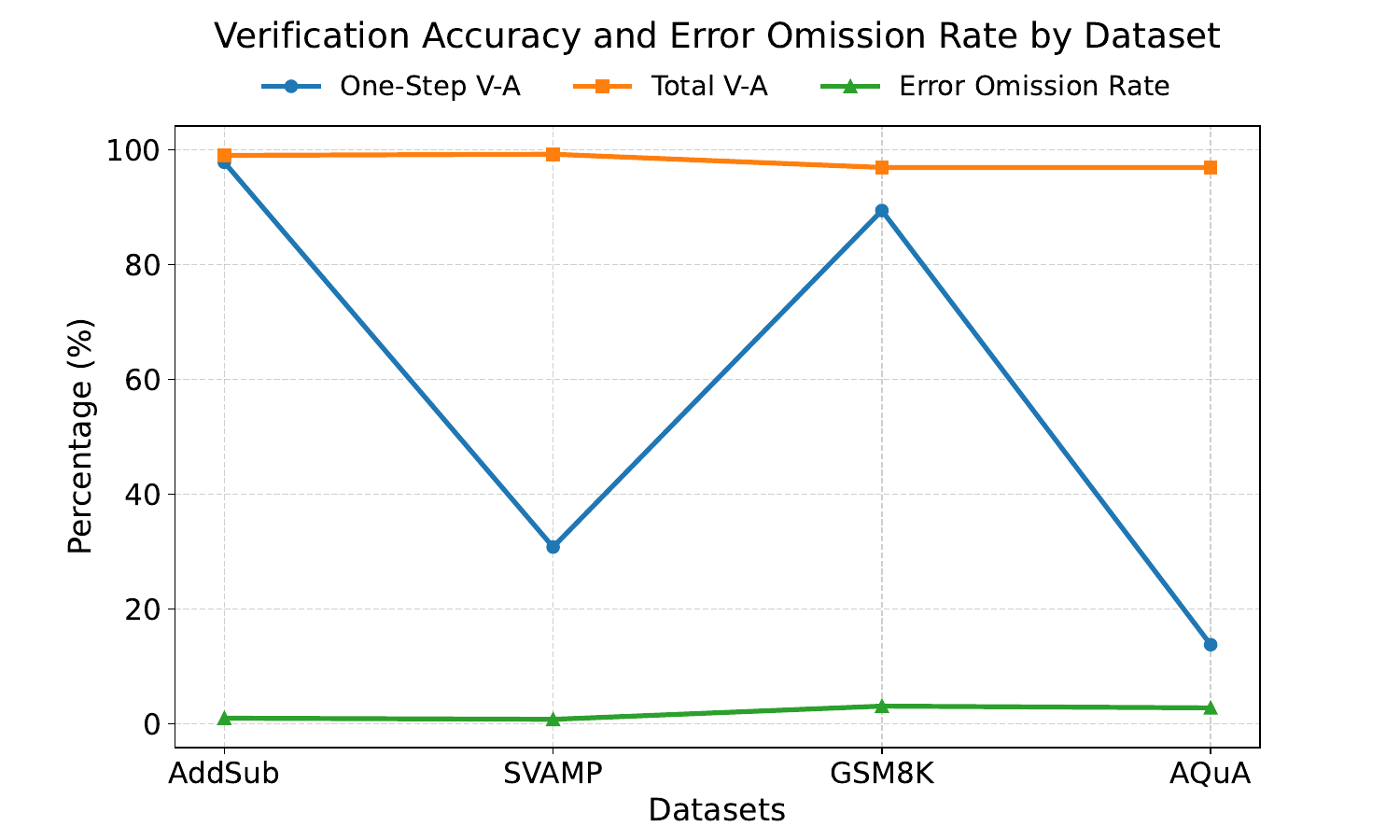}
    \caption{}
    \label{fig:accuracy}
  \end{subfigure}
  \hfill
  \begin{subfigure}{0.49\linewidth}
    \centering
    \includegraphics[width=\linewidth]{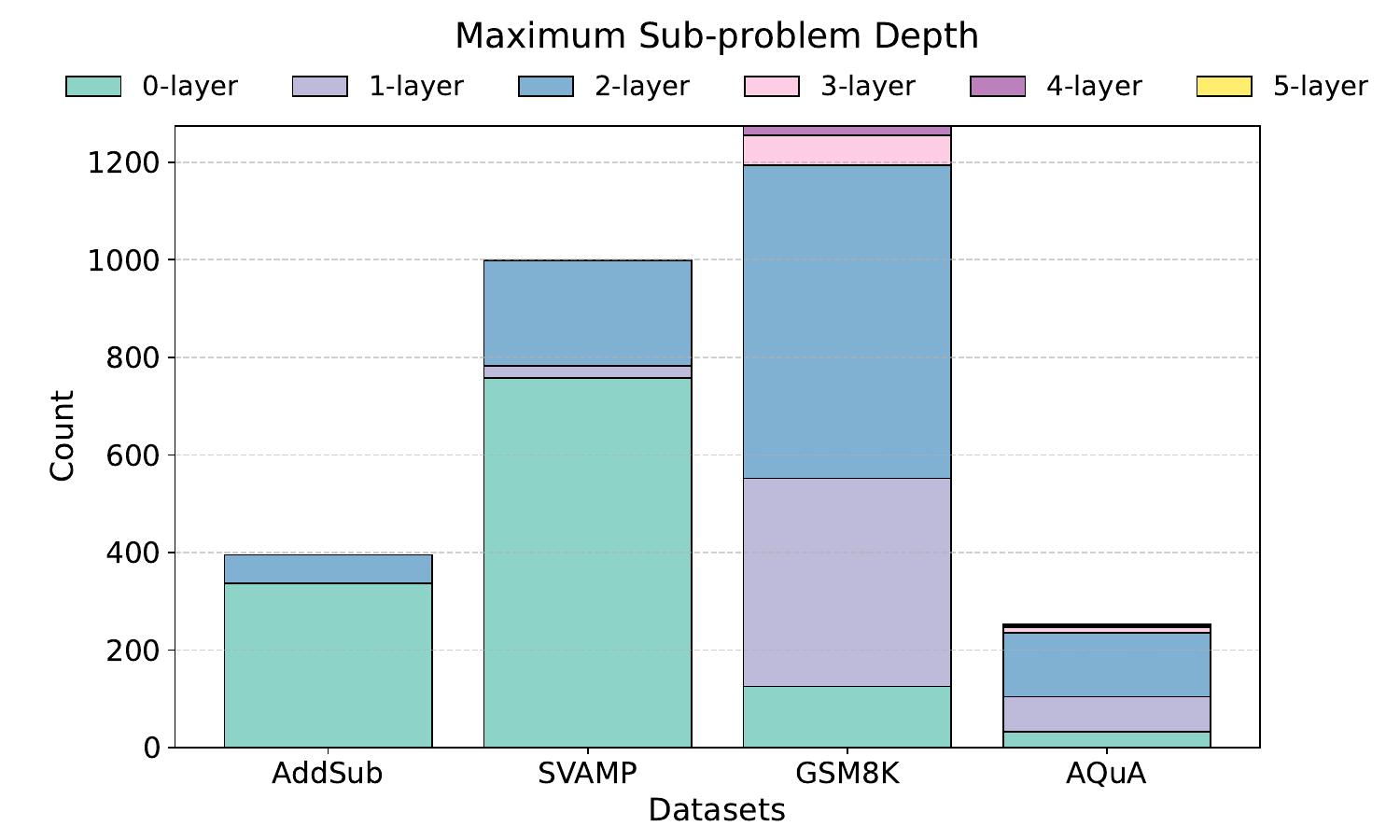}
    \caption{}
    \label{fig:max_layer}
  \end{subfigure}
  \caption{We present a visualization of the dataset analysis. Figure (a) shows the various accuracy metrics on the dataset, where V-A denotes Verification Accuracy. Figure (b) shows the maximum number of layers in the multi-layer Markov chains used in the dataset.}
  \label{fig:combined}
\end{figure*}

\subsection{Verification Analysis}
To systematically evaluate our model’s verification capability on mathematical reasoning, we built verification datasets from four standard math QA benchmarks: AddSub, SVAMP, GSM8K, and AQuA (see Figure~\ref{fig:combined}). Error samples are identified through strict comparison between generated final answers and Ground Truth (GT), with any discrepancy flagged as incorrect. Second, backward verification operates exclusively on the model's own generated reasoning trajectory, ensuring self-correction is fully grounded in internal context. Finally, the maximum layer count is dynamically controlled by the problem's maximum dependency number. We analyze three key metrics:
\begin{itemize}
\item One-Step Verification Accuracy: the fraction of problems whose initial solution is both correct and verified on the first attempt.
\item Total Verification Accuracy: the proportion of problems that are either correctly solved or whose errors are successfully detected.
\item Error Omission Rate: the fraction of incorrect solutions missed by the verifier.
\end{itemize}
One-Step Verification Accuracy is high on AddSub (97.8\%) and GSM8K (89.4\%), indicating strong self-verification on problems with simple or clear reasoning paths. It drops sharply on SVAMP (30.8\%) and AQuA (13.8\%), reflecting the difficulty for models to produce completely flawless and verifiable solutions in a single generation for semantically complex or multi-step reasoning problems. In contrast, Total Verification Accuracy remains consistently high across all datasets, demonstrating that CLoT reliably ensures output correctness by either producing correct answers or flagging errors. Correspondingly, the Error Omission Rate is low, confirming the verifier’s high recall. Notably, AQuA achieves an omission rate of only 2.8\%, comparable to GSM8K, despite its low one-step accuracy. This highlights the effectiveness of multi-round verification in catching errors even from poor initial solutions. Finally, the maximum layer counts are shown in Figure \ref{fig:max_layer}. The results indicate that AddSub and SVAMP have at most two layers of subproblems, whereas GSM8K and AQuA exhibit three or more layers, demonstrating that more challenging tasks require multiple rounds of the "generate-and-verify" cycle.
\begin{table}[t]
\centering
\renewcommand\arraystretch{1.0}
\begin{tabular}{lc}
\toprule
Model & Accuracy \\
\midrule
Qwen-2.5-1.5B & 75.33\% \\
Qwen-2.5-1.5B-CLoT-Instruct & 81.67\% \\
\bottomrule
\end{tabular}
\caption{SFT Performance Comparison}
\label{tab:sft_performance}
\end{table} 
\subsection{Datasets Analysis}
CLoT-Instruct utilizes instruction tuning to enhance models' self-verification capabilities. We fine-tuned Qwen-2.5-1.5B on CLoT-GSM8K-Instruct (2,700 training, 300 test samples). To mitigate trajectory inconsistency, we evaluated performance strictly via final answer matching, the results are shown in Table \ref{tab:sft_performance}. The fine-tuned model's GSM8K accuracy rose from 75.33\% to 81.67\% (+6.34\%), proving our data quality and method's efficacy. Future work will scale this pipeline to larger datasets and explore automated trajectory filtering to boost generalization and robustness in complex reasoning.

\begin{table*}
\centering
\tabcolsep=0.3cm
\renewcommand\arraystretch{1}
\small
\begin{tabular}{l|c|cccc|c}
\toprule
\textbf{Models} &\textbf{Token} & \textbf{AddSub} & \textbf{GSM8K} & \textbf{SVAMP} & \textbf{AQuA}  & \textbf{Avg.}\\
\midrule
CoT (baseline)&60k &94.9&90.9 &93.4&81.4&90.2\\
CoT + SV &98k &94.0 &93.5 &92.3&81.4&90.3\\
CoT + HMC&112k&94.9&90.9&93.4&81.0&90.1\\
CoT + RHMC&325k&94.4&93.5&93.9&84.6&91.6\\
CoT + RHMC + HP&136k&\ccr{99.0}&\ccr{94.6}&\ccr{94.9}&\ccr{85.8}&\ccr{93.6}\\
\bottomrule
\end{tabular}
\caption{Ablation study results of CLoT based on gpt-4o-mini. The experiment used four datasets and token consumption to evaluate the contributions of three components. The token consumption corresponds to the total number of tokens used to solve the same 100 GSM8K problems (idx0 – idx99).}
\label{ablation}
\end{table*}
\subsection{Ablation Analysis}
To systematically evaluate the effectiveness of each component in our proposed method, we conduct a series of ablation studies using the gpt-4o-mini model. As shown in Table~\ref{ablation}, we incrementally introduce four key designs: (1) standard Chain-of-Thought (CoT) with a single direct answer verification (CoT + Self-Verifying, or CoT + SV), serving as the baseline; (2) CoT enhanced with Hierarchical Markov Chain decomposition (CoT + HMC) without any verification, to assess the benefit of structured reasoning alone; (3) CoT with Reversible Hierarchical Markov Chain (CoT + RHMC), which adds step-by-step verification at each reasoning level to improve overall reliability; and (4) CoT + RHMC further augmented with Hierarchical Pruning (HP), where lower-level verification steps are skipped once all subproblems at a higher level are confirmed correct.

The results indicate that CoT+SV yields only a marginal improvement over the baseline CoT. This highlights the limitations of unstructured verification, which often struggles with complex error localization. When upgrading to CoT+RHMC, the average accuracy significantly improves to 91.6\%. However, this method generates excessively long reasoning chains (325k tokens), requiring the model to meticulously verify every low-level sub-problem even when the high-level logic is already sound. Finally, CoT+RHMC+HP further boosts the average accuracy to 93.6\% while drastically reducing token consumption to 136k. Although pruning is fundamentally an efficiency mechanism, its accuracy gains stem from the attention characteristics of LLMs. Excessively long reasoning chains cause attention dilution, whereas HP maintains concise reasoning chains by pruning redundant checks after macroscopic consistency is verified, thereby preventing the model from over-correcting valid reasoning steps.
\section{Conclusion}
In this paper, we introduced Cognitive Loop of Thought (CLoT), a novel reasoning framework that addresses the inherent "memorylessness" and error propagation issues in traditional Chain-of-Thought (CoT) prompting. By modeling the reasoning process as a Reversible Hierarchical Markov Chain (RHMC), we bridge the gap between sequential token generation and human-like cognitive verification. Our core contribution, the backward verification mechanism, enables LLMs to treat reasoning as a closed loop, where conclusions are validated by reconstructing original premises. To ensure this process remains computationally viable, we implemented a hierarchical pruning strategy that leverages the dependency structure of sub-problems to bypass redundant checks. Our experiments across six benchmarks demonstrate that CLoT consistently achieves state-of-the-art performance, notably reaching 99.0\% accuracy on the AddSub dataset. Furthermore, our ablation studies confirm that the synergy between reversible logic and hierarchical pruning allows for a 41.8\% reduction in token overhead without sacrificing reasoning integrity. The release of CLoT-Instruct provides a foundational resource for the community to further explore bidirectional instruction tuning. We believe that shifting from purely forward-moving chains to cognitive loops is a critical step toward developing more reliable, self-correcting, and autonomous reasoning agents.

\section*{Limitations}
This study has two main limitations.
First, CLoT relies on the model's intrinsic ability to perform backward reasoning. Our experiments show that while large-scale models like gpt-4o excel at this, smaller or less capable models may struggle with the logical inversion required for effective self-verification, potentially leading to false negatives during the pruning process. Second, our current evaluation focuses primarily on mathematical, logical, and commonsense reasoning tasks where conclusions are relatively deterministic. In more subjective or open-ended tasks, such as creative writing or legal argumentation, defining a clear backward verification logic is more challenging, as there may not be a unique set of premises that lead to a specific conclusion.



\bibliography{custom}

@article{wang2023shepherd,
  title={Shepherd: A critic for language model generation},
  author={Wang, Tianlu and Yu, Ping and et al.},
  journal={arXiv preprint arXiv:2308.04592},
  year={2023}
}

@inproceedings{weng-etal-2023-large,
    title = "Large Language Models are Better Reasoners with Self-Verification",
    author = "Weng, Yixuan  and
      Zhu, Minjun  and
      Xia, Fei  and et al.",
    booktitle = "Findings of the Association for Computational Linguistics: EMNLP 2023",
    url = "https://aclanthology.org/2023.findings-emnlp.167/",
    doi = "10.18653/v1/2023.findings-emnlp.167",
    pages = "2550--2575",
}

@article{li2024confidence,
  title={Confidence matters: Revisiting intrinsic self-correction capabilities of large language models},
  author={Li, Loka and Chen, Zhenhao and Chen, Guangyi and et al.},
  journal={arXiv preprint arXiv:2402.12563},
  year={2024}
}

@inproceedings{
zheng2024progressivehint,
title={Progressive-Hint Prompting Improves Reasoning in Large Language Models},
author={Chuanyang Zheng and Zhengying Liu and Enze Xie and et al.},
booktitle={AI for Math Workshop @ ICML 2024},
year={2024},
url={https://openreview.net/forum?id=UkFEs3ciz8}
}

@inproceedings{10.5555/3692070.3692341,
author = {Chen, Sijia and Li, Baochun},
title = {Toward adaptive reasoning in large language models with thought rollback},
year = {2024},
booktitle = {Proceedings of the 41st International Conference on Machine Learning},
articleno = {271},
numpages = {24},
}

@article{yang2024confidence,
  title={Confidence vs critique: A decomposition of self-correction capability for llms},
  author={Yang, Zhe and Zhang, Yichang and Wang, Yudong and et al.},
  journal={arXiv preprint arXiv:2412.19513},
  year={2024}
}

@article{achiam2023gpt,
  title={Gpt-4 technical report},
  author={Achiam, Josh and Adler, Steven and Agarwal, Sandhini and et al.},
  journal={arXiv preprint arXiv:2303.08774},
  year={2023}
}

@article{chowdhery2023palm,
  title={Palm: Scaling language modeling with pathways},
  author={Chowdhery, Aakanksha and Narang, Sharan and Devlin, Jacob and et al.},
  journal={Journal of Machine Learning Research},
  volume={24},
  number={240},
  pages={1--113},
  year={2023}
}

@article{touvron2023llama,
  title={Llama: Open and efficient foundation language models},
  author={Touvron, Hugo and Lavril, Thibaut and Izacard, Gautier and et al.},
  journal={arXiv preprint arXiv:2302.13971},
  year={2023}
}

@inproceedings{huang-etal-2023-large,
    title = "Large Language Models Can Self-Improve",
    author = "Huang, Jiaxin  and
      Gu, Shixiang  and
      Hou, Le  and
      et al.",
    booktitle = "Proceedings of the 2023 Conference on Empirical Methods in Natural Language Processing",
    year = "2023",
    url = "https://aclanthology.org/2023.emnlp-main.67/",
    doi = "10.18653/v1/2023.emnlp-main.67",
    pages = "1051--1068",
}

@article{zhao2023survey,
  title={A survey of large language models},
  author={Zhao, Wayne Xin and Zhou, Kun and Li, Junyi and et al.},
  journal={arXiv preprint arXiv:2303.18223},
  volume={1},
  number={2},
  year={2023}
}

@article{zhou2024larger,
  title={Larger and more instructable language models become less reliable},
  author={Zhou, Lexin and Schellaert, Wout and Mart{\'\i}nez-Plumed, Fernando and et al.},
  journal={Nature},
  volume={634},
  number={8032},
  pages={61--68},
  year={2024},
  publisher={Nature Publishing Group UK London}
}

@inproceedings{NEURIPS2022_8bb0d291,
 author = {Kojima, Takeshi and Gu, Shixiang (Shane) and Reid, Machel and et al.},
 booktitle = {Advances in Neural Information Processing Systems},
 pages = {22199--22213},
 title = {Large Language Models are Zero-Shot Reasoners},
 url = {https://proceedings.neurips.cc/paper_files/paper/2022/file/8bb0d291acd4acf06ef112099c16f326-Paper-Conference.pdf},
 volume = {35},
 year = {2022}
}

@article{10.1162/tacl_a_00713,
    author = {Kamoi, Ryo and Zhang, Yusen and Zhang, Nan and et al.},
    title = {When Can LLMs Actually Correct Their Own Mistakes? A Critical Survey of Self-Correction of LLMs},
    journal = {Transactions of the Association for Computational Linguistics},
    volume = {12},
    pages = {1417-1440},
    year = {2024},
    month = {11},
    issn = {2307-387X},
    doi = {10.1162/tacl_a_00713},
    url = {https://doi.org/10.1162/tacl_a_00713},
}

@article{10.1162/tacl_a_00660,
    author = {Pan, Liangming and Saxon, Michael and Xu, Wenda and et al.},
    title = {Automatically Correcting Large Language Models: Surveying the Landscape of Diverse Automated Correction Strategies},
    journal = {Transactions of the Association for Computational Linguistics},
    volume = {12},
    pages = {484-506},
    year = {2024},
    month = {05},
    issn = {2307-387X},
    doi = {10.1162/tacl_a_00660},
    url = {https://doi.org/10.1162/tacl_a_00660},
}

@inproceedings{
stechly2023gpt,
title={{GPT}-4 Doesn{\textquoteright}t Know It{\textquoteright}s Wrong: An Analysis of Iterative Prompting for Reasoning Problems},
author={Kaya Stechly and Matthew Marquez and Subbarao Kambhampati},
booktitle={NeurIPS 2023 Foundation Models for Decision Making Workshop},
year={2023},
url={https://openreview.net/forum?id=PMtZjDYB68}
}

@inproceedings{tyen-etal-2024-llms,
    title = "{LLM}s cannot find reasoning errors, but can correct them given the error location",
    author = "Tyen, Gladys  and
      Mansoor, Hassan  and
      Carbune, Victor  and
      et al.",
    year = "2024",
    publisher = "Association for Computational Linguistics",
    url = "https://aclanthology.org/2024.findings-acl.826/",
    doi = "10.18653/v1/2024.findings-acl.826",
    pages = "13894--13908",
}

@article{jiang2024self,
  title={Self-[in] correct: Llms struggle with refining self-generated responses},
  author={Jiang, Dongwei and Zhang, Jingyu and Weller, Orion and et al.},
  journal={CoRR},
  year={2024}
}

@inproceedings{NEURIPS2022_9d560961,
 author = {Wei, Jason and Wang, Xuezhi and Schuurmans, Dale and et al.},
 booktitle = {Advances in Neural Information Processing Systems},
 pages = {24824--24837},
 title = {Chain-of-Thought Prompting Elicits Reasoning in Large Language Models},
 url = {https://proceedings.neurips.cc/paper_files/paper/2022/file/9d5609613524ecf4f15af0f7b31abca4-Paper-Conference.pdf},
 volume = {35},
 year = {2022}
}

@inproceedings{
wang2023selfconsistency,
title={Self-Consistency Improves Chain of Thought Reasoning in Language Models},
author={Xuezhi Wang and Jason Wei and Dale Schuurmans and et al.},
booktitle={The Eleventh International Conference on Learning Representations },
year={2023},
url={https://openreview.net/forum?id=1PL1NIMMrw}
}

@inproceedings{
zhang2023automatic,
title={Automatic Chain of Thought Prompting in Large Language Models},
author={Zhuosheng Zhang and Aston Zhang and Mu Li and Alex Smola},
booktitle={The Eleventh International Conference on Learning Representations },
year={2023},
url={https://openreview.net/forum?id=5NTt8GFjUHkr}
}

@inproceedings{
fu2023complexitybased,
title={Complexity-Based Prompting for Multi-step Reasoning},
author={Yao Fu and Hao Peng and Ashish Sabharwal and et al.},
booktitle={The Eleventh International Conference on Learning Representations },
year={2023},
url={https://openreview.net/forum?id=yf1icZHC-l9}
}

@inproceedings{NEURIPS2023_271db992,
 author = {Yao, Shunyu and Yu, Dian and Zhao, Jeffrey and et al.},
 booktitle = {Advances in Neural Information Processing Systems},
 pages = {11809--11822},
 title = {Tree of Thoughts: Deliberate Problem Solving with Large Language Models},
 url = {https://proceedings.neurips.cc/paper_files/paper/2023/file/271db9922b8d1f4dd7aaef84ed5ac703-Paper-Conference.pdf},
 volume = {36},
 year = {2023}
}

@inproceedings{besta2024graph,
  title={Graph of thoughts: Solving elaborate problems with large language models},
  author={Besta, Maciej and Blach, Nils and Kubicek, Ales and et al.},
  booktitle={Proceedings of the AAAI conference on artificial intelligence},
  volume={38},
  number={16},
  pages={17682--17690},
  year={2024}
}

@article{teng2025atom,
  title={Atom of thoughts for markov llm test-time scaling},
  author={Teng, Fengwei and Yu, Zhaoyang and Shi, Quan and et al.},
  journal={arXiv preprint arXiv:2502.12018},
  year={2025}
}

@article{zhu2025understanding,
  title={Understanding before reasoning: Enhancing chain-of-thought with iterative summarization pre-prompting},
  author={Zhu, Dong-Hai and Xiong, Yu-Jie and Zhang, Jia-Chen and et al.},
  journal={arXiv preprint arXiv:2501.04341},
  year={2025}
}

@article{cobbe2021training,
  title={Training verifiers to solve math word problems},
  author={Cobbe, Karl and Kosaraju, Vineet and Bavarian, Mohammad and et al.},
  journal={arXiv preprint arXiv:2110.14168},
  year={2021}
}

@inproceedings{
hendrycks2021measuring,
title={Measuring Mathematical Problem Solving With the {MATH} Dataset},
author={Dan Hendrycks and Collin Burns and Saurav Kadavath and et al.},
booktitle={Thirty-fifth Conference on Neural Information Processing Systems Datasets and Benchmarks Track (Round 2)},
year={2021},
url={https://openreview.net/forum?id=7Bywt2mQsCe}
}

@inproceedings{hosseini2014learning,
  title={Learning to solve arithmetic word problems with verb categorization},
  author={Hosseini, Mohammad Javad and Hajishirzi, Hannaneh and Etzioni, Oren and Kushman, Nate},
  booktitle={Proceedings of the 2014 conference on empirical methods in natural language processing (EMNLP)},
  pages={523--533},
  year={2014}
}

@inproceedings{patel-etal-2021-nlp,
    title = "Are {NLP} Models really able to Solve Simple Math Word Problems?",
    author = "Patel, Arkil  and
      Bhattamishra, Satwik  and
      Goyal, Navin",
    booktitle = "Proceedings of the 2021 Conference of the North American Chapter of the Association for Computational Linguistics: Human Language Technologies",
    year = "2021",
    url = "https://aclanthology.org/2021.naacl-main.168/",
    doi = "10.18653/v1/2021.naacl-main.168",
    pages = "2080--2094",
}

@inproceedings{ling-etal-2017-program,
    title = "Program Induction by Rationale Generation: Learning to Solve and Explain Algebraic Word Problems",
    author = "Ling, Wang  and
      Yogatama, Dani  and
      Dyer, Chris  and
      Blunsom, Phil",
    booktitle = "Proceedings of the 55th Annual Meeting of the Association for Computational Linguistics (Volume 1: Long Papers)",
    year = "2017",
    url = "https://aclanthology.org/P17-1015/",
    doi = "10.18653/v1/P17-1015",
    pages = "158--167",
}

@inproceedings{talmor-etal-2019-commonsenseqa,
    title = "{C}ommonsense{QA}: A Question Answering Challenge Targeting Commonsense Knowledge",
    author = "Talmor, Alon  and
      Herzig, Jonathan  and
      Lourie, Nicholas  and
      Berant, Jonathan",
    booktitle = "Proceedings of the 2019 Conference of the North {A}merican Chapter of the Association for Computational Linguistics: Human Language Technologies, Volume 1 (Long and Short Papers)",
    year = "2019",
    url = "https://aclanthology.org/N19-1421/",
    doi = "10.18653/v1/N19-1421",
    pages = "4149--4158",
}

@inproceedings{
yasunaga2024large,
title={Large Language Models as Analogical Reasoners},
author={Michihiro Yasunaga and Xinyun Chen and Yujia Li and et al.},
booktitle={The Twelfth International Conference on Learning Representations},
year={2024},
url={https://openreview.net/forum?id=AgDICX1h50}
}

@inproceedings{jiang-etal-2024-forward,
    title = "Forward-Backward Reasoning in Large Language Models for Mathematical Verification",
    author = "Jiang, Weisen  and
      Shi, Han  and
      Yu, Longhui  and
      et al.",
    booktitle = "Findings of the Association for Computational Linguistics: ACL 2024",
    month = aug,
    year = "2024",
    address = "Bangkok, Thailand",
    publisher = "Association for Computational Linguistics",
    url = "https://aclanthology.org/2024.findings-acl.397/",
    doi = "10.18653/v1/2024.findings-acl.397",
    pages = "6647--6661",
}

@inproceedings{yang2025markov,
    title = "{M}arkov Chain of Thought for Efficient Mathematical Reasoning",
    author = "Yang, Wen  and
      Liao, Minpeng  and
      Fan, Kai",
    booktitle = "Proceedings of the 2025 Conference of the Nations of the Americas Chapter of the Association for Computational Linguistics: Human Language Technologies (Volume 1: Long Papers)",
    month = apr,
    year = "2025",
    url = "https://aclanthology.org/2025.naacl-long.365/",
    doi = "10.18653/v1/2025.naacl-long.365",
    pages = "7132--7157",
    ISBN = "979-8-89176-189-6"
}
\clearpage
\appendix
\section{datasets}
\subsection{Mathematical Reasoning Datasets}
We evaluate CLoT on four mathematical QA datasets and one multiple-choice dataset:

1) MATH \cite{hendrycks2021measuring}\& SVAMP \cite{patel-etal-2021-nlp}: Given the extensive size of these benchmarks, we follow the established data processing procedure from prior research \cite{10.5555/3692070.3692341}, specifically evaluating the first 900 samples of MATH and the first 1,000 samples of SVAMP to ensure consistent comparison.

2) GSM8K \cite{cobbe2021training}\& AddSub \cite{hosseini2014learning}: We utilize the full test sets (1,319 and 395 samples, respectively) to assess the model's ability to handle grade-school level multi-step arithmetic word problems.

3) AQuA \cite{ling-etal-2017-program}: A mathematical multiple-choice dataset requiring algebraic reasoning. This helps evaluate CLoT's performance when the solution space is constrained by predefined options.
\subsection{Commonsense Reasoning Dataset}
CommonsenseQA (CSQA) \cite{talmor-etal-2019-commonsenseqa}: We use 1,220 samples from CSQA to test the framework's ability to leverage world knowledge and semantic relationships.


\begin{table}[h]
\centering
\tabcolsep=0.1cm
\renewcommand\arraystretch{1.2}
\small
\begin{tabular}{lccc}
\toprule
\textbf{Dataset} & \textbf{Reasoning Task} & \textbf{Answer Type} & \textbf{Number} \\ \midrule
MATH         & Mathematical  & Number   & 900    \\
GSM8K        & Mathematical   & Number  & 1,319  \\
AddSub       & Mathematical  & Number   & 395    \\
SVAMP        & Mathematical  & Number  & 1,000  \\
AQuA        & Mathematical   & Multi-choice & 254    \\
CSQA  & Commonsense      & Multi-choice  & 1,220  \\ \bottomrule
\end{tabular}
\caption{Overview of datasets utilized in CLoT experiments.}
\label{tab:CLoT_datasets}
\end{table}

\subsection{License}
All datasets used in this study are publicly available resources employed strictly for non-commercial research purposes. MATH~\cite{hendrycks2021measuring} and SVAMP~\cite{patel-etal-2021-nlp} are both released under the MIT License. GSM8K~\cite{cobbe2021training} is made available by OpenAI under the MIT License, while AQuA~\cite{ling-etal-2017-program} is distributed under the Apache License 2.0. AddSub~\cite{hosseini2014learning} and CommonsenseQA (CSQA)~\cite{talmor-etal-2019-commonsenseqa} do not specify explicit open-source licenses; both are publicly accessible and widely adopted for academic research in the NLP community.

\begin{table*}[ht]
\centering
\renewcommand\arraystretch{1.5}
\begin{tabular}{|l|p{8.8cm}|}
\hline
\textbf{Task} & \textbf{Answer-format Instructions} \\
\hline
GSM8K, SVAMP, AddSub, math & You can freely reason in your response, but please enclose the final answer within $<$answer$><$/answer$>$tags(pure number without units and explanations) \\
\hline
AQuA, CommonsenseQA & You can freely reason in your response, but please enclose the final option within $<$answer$><$/answer$>$tags(pure uppercase option without explanations)  \\
\hline
\end{tabular}
\caption{Answer-format instructions for different tasks.}
\label{tab:answer_formats}
\end{table*}

\section{Prompt Cases}
To support reproducible research, the full prompt templates utilized in our experimental pipeline are publicly released.
\begin{tcolorbox}[
  colback=gray!10,      
  colframe=black,    
  arc=1mm,               
  boxrule=0.5mm,            
  left=6pt,            
  right=6pt,             
  top=6pt,              
  bottom=6pt,
  title=\textbf{Initial Reasoning},
  before skip=6pt,   
  after skip=6pt,     
  breakable
]
\textbf{System Prompt:} \\
You are a precise math problem solver. Solve the given math problem step by step:\\

Please convert fractional answers to decimal form.\\
Please ensure that a final numerical answer is obtained.\\
Please extend your chain of thought as much as possible; the longer the chain of thought, the better.\\
        
You can freely reason in your response, but please enclose the final answer within <answer></answer> tags (pure number without units and explanations)\\

\textbf{Question:}(...)\\

\textbf{Response:}\\
<think>(...)</think>\\
<answer>(...)</answer>
\end{tcolorbox}

\begin{tcolorbox}[
  colback=gray!10,      
  colframe=black,    
  arc=1mm,               
  boxrule=0.5mm,            
  left=6pt,            
  right=6pt,             
  top=6pt,              
  bottom=6pt,
  title=\textbf{Problem Decomposition},
  before skip=6pt,   
  after skip=6pt,     
  breakable
]
\textbf{System Prompt:} \\
You are tasked with breaking down a math problem reasoning process into sub-questions. Please ensure that there are no repeated questions.\\

Instructions:\\
    1. Break down the reasoning process into a series of sub-questions\\
    2. Each sub-question should:\\
        - Do not ask for equation\\
        - Be written in interrogative form\\
        - Have a clear numerical answer\\
        - List its other sub-questions' indexes it depends (0-based, can be an empty list)\\
    3. Dependencies are defined as information needed to answer the current sub-question that:\\
        - Does NOT come directly from the original question\\
        - MUST come from the answers of previous sub-questions\\

formatter = """\\
    Format your response as the following JSON object:\\
    \{\{\\
        "sub-questions": [\\
            \{\{\\
                "description": "<clear interrogative question>",\\
                "answer": <numerical value without units>,\\
                "depend": [<indices of prerequisite sub-questions>]\\
            \}\},\\
            ...\\
        ],\\
        "answer": {answer}\\
    \}\}\\
\textbf{Question:}(...)\\
\textbf{Chain of Thought:}(...)\\

\textbf{Response:}\\
<think>(...)</think>\\
    \{\{\\
        "sub-questions": [\\
            \{\{\\
                "description": "(...)",\\
                "answer": (...),\\
                "depend": [(...)]\\
            \}\},\\
            ...\\
        ],\\
        "answer": {answer}\\
    \}\}\\
\end{tcolorbox}
\begin{tcolorbox}[
  colback=gray!10,      
  colframe=black,    
  arc=1mm,               
  boxrule=0.5mm,            
  left=6pt,            
  right=6pt,             
  top=6pt,              
  bottom=6pt,
  title=\textbf{Problem Reconstruction},
  before skip=6pt,   
  after skip=6pt,     
  breakable
]
\textbf{System Prompt:} \\
Please replace one numerical value in the original problem with x. This value will be used to construct a verification question that checks the correctness of the original problem.\\
Only one numerical value should be replaced with x; all other information must remain unchanged.
Whenever possible, avoid using fractions or decimals when introducing x. \\
Please enclose the modified question within <question></question> tags and enclose The original value that was replaced by x (only one) within <answer></answer> tags (pure number without units and explanations)\\

\textbf{Question:}(...)\\

\textbf{Response:}\\
<question>(...)</question>\\
<answer>(...)</answer>
\end{tcolorbox}

\begin{tcolorbox}[
  colback=gray!10,      
  colframe=black,    
  arc=1mm,               
  boxrule=0.5mm,            
  left=6pt,            
  right=6pt,             
  top=6pt,              
  bottom=6pt,
  title=\textbf{Supplementary Information},
  before skip=6pt,   
  after skip=6pt,     
  breakable
]
\textbf{System Prompt:} \\
Please add the necessary information to the sub-question so that it can be answered independently of the original question.\\

Instructions:\\
    1. Search for information in the original question that is relevant to solving the sub-question.\\
    2. Combine the information with the sub-question in a logical way to form a new question that can be solved independently.\\
    3. If the sub-question is consistent with the original question, directly output the original question as the new question.\\

You are not required to solve the problem, but please enclose the new problem within <problem></problem> tags.\\

\textbf{Question:}(...)\\

\textbf{Response:}\\
<problem>(...)</problem>

\end{tcolorbox}

\begin{tcolorbox}[
  colback=gray!10,      
  colframe=black,    
  arc=1mm,               
  boxrule=0.5mm,            
  left=6pt,            
  right=6pt,             
  top=6pt,              
  bottom=6pt,
  title=\textbf{Conclusion},
  before skip=6pt,   
  after skip=6pt,     
  breakable
]
\textbf{System Prompt:} \\
You are given a main question and a related sub-question. Your task is to analyze how the sub-question contributes to solving the main question, and to summarize the key strategies or patterns used in correctly answering the sub-question. Focus on identifying problem-solving techniques and any dependencies between the sub-question and the final solution.\\

Please enclose the problem-solving techniques within <solve></solve> tags.\\

\textbf{Question:}(...)\\
\textbf{Sub-question:}(...)\\

\textbf{Response:}\\
<solve>(...)</solve>

\end{tcolorbox}
\begin{tcolorbox}[
  colback=gray!10,      
  colframe=black,    
  arc=1mm,               
  boxrule=0.5mm,            
  left=6pt,            
  right=6pt,             
  top=6pt,              
  bottom=6pt,
  title=\textbf{Final Reasoning},
  before skip=6pt,   
  after skip=6pt,     
  breakable
]
\textbf{System Prompt:} \\
You are a precise math problem solver. Solve the given math problem step by step. Please refer to the following sub-questions, sub-answers and problem-solving experiences.\\

Please convert fractional answers to decimal form.\\
Please ensure that a final numerical answer is obtained.\\

You can freely reason in your response, but please enclose the final answer within <answer></answer> tags (pure number without units and explanations)\\

\textbf{Question:}(...)\\
\textbf{Sub-question:}(...)\\

\textbf{Response:}\\
<think>(...)</think>\\
<answer>(...)</answer>

\end{tcolorbox}

\section{Answer Formation}
The following section illustrates the multi-stage reasoning process of CLoT, where discrete steps are facilitated through Chain-of-Thought (CoT) prompting. Specifically, Tables \ref{tab:math} and \ref{tab:math1} exemplify the detailed reasoning trajectories for mathematical problems. To ensure consistency and facilitate evaluation, the final output generation phase adheres to the standardized formats presented in Table \ref{tab:answer_formats}.

\begin{table}[h]
\centering
\renewcommand\arraystretch{1.2}
\begin{tabular}{|p{1\linewidth}|}
\hline
\textbf{SVAMP Problem}\\
\hline
Q: Kylar went to the store to buy glasses for his new apartment. One glass costs \$5, but every second glass costs only 60\% of the price. Kylar wants to buy 16 glasses. How much does he need to pay?\\
\hline
\textbf{Step 1}, First CoT \\
\hline
Q: Let's think step by step.\\
***Thought***
$<$answer$>64<$/answer$>$\\
\hline
\textbf{Step 2}, Backward Verification\\
\hline
Q: Please replace one numerical value in the original problem with X and incorporate the answer as a known condition.\\
Kylar went to the store to buy glasses for his new apartment. One glass costs X, but every second glass costs only 60\% of the price. Kylar wants to buy 16 glasses. Knowing that he need to pay  \$64 for them. What's the X?\\
\hline
\textbf{Step 3}, Solve the new problem\\
\hline
Q: Let's think step by step.\\
***Thought***
$<$answer$>5<$/answer$>$\\
\hline
\textbf{Step 4}, Judge\\
\hline
5 $=$ 5\\
\hline
\textbf{Answer}: 64\\
\hline
\end{tabular}
\caption{Examples of Successful Verification by CLoT on Mathematical Problem Datasets.}
\label{tab:math}
\end{table}

\begin{table}[h]
\centering
\begin{tabular}{lc}
\toprule
\textbf{Method} & \textbf{Accuracy (\%)} \\
\midrule
CoT & 26.67 \\
CoT-SC & 23.33 \\
CLoT (Ours) & 30.00 \\
\bottomrule
\end{tabular}
\caption{Performance comparison on the AIME dataset (gpt-4).}
\label{tab:aime}
\end{table}

\begin{table*}[t]
\centering
\renewcommand\arraystretch{1.5}
\begin{tabular}{|p{1\linewidth}|}
\hline
\textbf{SVAMP Problem}\\
\hline
Q: Kylar went to the store to buy glasses for his new apartment. One glass costs \$5, but every second glass costs only 60\% of the price. Kylar wants to buy 16 glasses. How much does he need to pay?\\
\hline
\textbf{Step 1}, First CoT \\
\hline
Q: Let's think step by step.\\
***Thought***
$<$answer$>96<$/answer$>$\\
\hline
\textbf{Step 2}, Backward Verification\\
\hline
Q: Please replace one numerical value in the original problem with X and incorporate the answer as a known condition.\\
Kylar went to the store to buy glasses for his new apartment. One glass costs X, but every second glass costs only 60\% of the price. Kylar wants to buy 16 glasses. Knowing that he need to pay  \$64 for them. What's the X?\\
\hline
\textbf{Step 3}, Solve the new problem\\
\hline
Q: Let's think step by step.\\
***Thought***
$<$answer$>10<$/answer$>$\\
\hline
\textbf{Step 4}, Judge\\
\hline
4 $\neq$ 5\\
\hline
\textbf{Step 5}, Decompose sub-problems\\
\hline
Q: You are tasked with breaking down a math problem reasoning process into sub-problems.\\
q1, q2, q3,...\\
\hline
\textbf{Step 6}, For each subproblem, iteratively execute Steps 1–4.\\
\hline
If all subproblems at the same level pass verification, exit the loop.\\
\hline
\textbf{Step 7}, Second CoT\\
\hline
Q: Let's think step by step. Please refer to the following sub-problems and sub-answers.\\
***Thought***
$<$answer$>64<$/answer$>$\\
\hline
\textbf{Answer}: 64\\
\hline
\end{tabular}
\caption{Examples of CLoT's verification failures on mathematical problem datasets.}
\label{tab:math1}
\end{table*}

\section{AIME Result}
We have supplemented our experiments with the AIME using gpt-4. The results, summarized in Table \ref{tab:aime}, show that CLoT (30.00\%) outperforms both CoT (26.67\%) and CoT-SC (23.33\%). Notably, on high-difficulty tasks, the performance of CoT-SC degrades, whereas CLoT continues to improve. This demonstrates that our verification mechanism is more effective than simple self-consistency sampling in complex reasoning scenarios.

\section{Declaration of AI Assistance}
In the preparation of this manuscript, AI-assisted tools were used exclusively for language polishing, grammar correction, and stylistic refinement of the text. These tools were not employed in the conception or design of the research methodology, the development of algorithms, the implementation of code, the analysis of experimental results, or the formulation of core scientific claims.

\section{Algorithm}
This section formalizes the operational workflow of the \textbf{Cognitive Loop of Thought} (CLoT) framework in Algorithm \ref{alg:CLoT_Reasoning}. CLoT achieves robust multi-step reasoning through hierarchical state transitions that decompose problems into layered Markov chains, coupled with a reversible verification mechanism that validates logical validity via backward justification and eliminates redundant computation through consistency-aware hierarchical pruning. These mechanisms are theoretically grounded in Reversible Hierarchical Markov Chains (RHMC).
\begin{algorithm}[t]
\caption{CLoT Reasoning}
\label{alg:CLoT_Reasoning}

\SetKwInput{KwIn}{Input}
\SetKwInput{KwOut}{Output}

\KwIn{Original problem $q_1^{(L)}$ at top layer $L$}
\KwOut{Final answer $a$ and global verification status $\mathcal{V}$}

\BlankLine
\tcp{Forward: top-down hierarchical deduction}
\For{$l = L$ \KwTo $1$}{
    $t \leftarrow 1$\;
    \While{not terminated}{
        $q_{t+1}^{(l)} \sim p(q_{t+1}^{(l)} \mid q_t^{(l)}, s_t^{(l)})$\;
        $t \leftarrow t+1$\;
    }
    $T_l \leftarrow t$\;
    \lIf{$l>1$}{$q_1^{(l-1)} \sim p(q_1^{(l-1)} \mid q_{1:T_l}^{(l)})$}
}
$a \leftarrow q_{T_1}^{(1)}$\;

\BlankLine
\tcp{Backward: top-down verification and pruning}
\For{$l = L$ \KwTo $1$}{
    $\mathcal{V} \leftarrow 1$\;
    \For{$t = T_l-1$ \KwTo $1$}{
        \lIf{$\text{Verify}(q_t^{(l)} \mid q_{t+1}^{(l)})$ fails}{$\mathcal{V} \leftarrow 0$; \textbf{break}}
    }
    \lIf{$\mathcal{V}=1$}{\textbf{break} \tcp*{Hierarchical pruning}}
}

\Return{$a, \mathcal{V}$}
\end{algorithm}

\section{Baseline}
Our baselines include classical prompting methods (CoT, \cite{NEURIPS2022_8bb0d291}), CoT with Self-Consistency (CoT-SC, n = 5, \cite{wang2023selfconsistency}), CoT with Iterative Summarization Pre-Prompting (ISP-2, \cite{zhu2025understanding}), Analogical Reasoning (AR, \cite{yasunaga2024large}), Complex-CoT (C-CoT, \cite{fu2023complexitybased}) and Thought Rollback (Tr, \cite{10.5555/3692070.3692341}).
\end{document}